# New metal-plastic hybrid additive manufacturing strategy: Fabrication of arbitrary metal-patterns on external and even internal surfaces of 3D plastic structures

Kewei Song[1], Yue Cui[1], Tiannan Tao[1], Xiangyi Meng[1], Michinari Sone[2], Masahiro Yoshino[2], Shinjiro Umezu[1,4] *, Hirotaka Sato[3,]*

[1] *Graduate School of Creative Science and Engineering, Department of Modern Mechanical Engineering, Waseda University, 3-4-1 Okubo, Shinjuku-ku, Tokyo 169-8555, Japan.*

[2] *Research and Development div., Yoshino Denka Kogyo, Inc., Japan.*

[3] *School of Mechanical and Aerospace Engineering, Nanyang Technological University, N3.2 – 01- 20, 65 Nanyang Drive, Singapore 637460, Singapore.*

[4] *Department of Modern Mechanical Engineering, Waseda University, 3-4-1 Okubo, Shinjuku-ku, Tokyo 169-8555, Japan.*

*Corresponding authors:

Shinjiro Umezu, Professor, E-mail: umeshin@waseda.jp

Hirotaka Sato, Associate Professor, E-mail: hirosato@ntu.edu.sg

**Abstract**

Constructing precise micro-nano metal patterns on complex three-dimensional (3D) plastic parts allows the fabrication of functional devices for advanced applications. However, this patterning is currently expensive and requires complex processes with long manufacturing lead time. The present work demonstrates a process for the fabrication of micro-nano 3D metal-plastic composite structures with arbitrarily complex shapes. In this approach, a light-cured resin is modified to prepare an active precursor capable of allowing subsequent electroless plating (ELP). A multi-material digital light processing 3D printer was newly developed to enable the fabrication of parts containing regions made of either standard resin or active precursor resin nested within each other. Selective 3D ELP processing of such parts provided various metal-plastic composite parts having complicated hollow micro-nano structures with specific topological relationships on a size scale as small as 40 μm. Using this technique, 3D metal topologies that cannot be manufactured by traditional methods are possible, and metal patterns can be produced inside plastic parts as a means of further miniaturizing electronic devices. The proposed method can also generate metal coatings exhibiting improved adhesion of metal to plastic substrate. Based on this technique, several sensors composed of different functional nonmetallic materials and specific metal patterns were designed and fabricated. The present results demonstrate the viability of the proposed method and suggest potential applications in the fields of smart 3D micro-nano electronics, 3D wearable devices, micro/nano-sensors, and health care.

## 1. Introduction

The formation of specific metal patterns on three-dimensional (3D) plastic parts has attracted a great deal of research interest because of potential applications in 3D smart electronics [1-3], telecommunication technology [4-6], micro/nano-sensors [7], microelectromechanical systems (MEMS) [8] and even quantum science [9]. In contrast to traditional two-dimensional (2D) printed circuit boards (PCBs) [10-12], 3D metal-plastic composite functional devices have more complex and sophisticated structures with higher degrees of design freedom and higher integration. The selective construction of metallized layouts in specific regions of 3D substrates can allow the fabrication of interconnecting devices having a variety of complicated geometric shapes. These 3D metal-plastic structures greatly reduce size requirements relative to standard planar printed electronics and thus permit the further miniaturization of functional devices. Even so, it is difficult to build interconnected metal patterns on the surfaces of more complex 3D parts using traditional microfabrication techniques such as lithography, deposition, etching and release. Because the manufacturing of 3D mental-plastic components requires laser direct structuring (LDS), the fabrication of such parts can be expensive and may involve long production cycle, a high degree of complexity and low design flexibility [13-19].

The combination of computer aided design (CAD) and 3D printing (3DP) allows the manufacture of a wide range of complex shapes via a layer-by-layer manufacturing process. Despite the greater design flexibility and superior processing capability of 3DP, this technology has not reached its full potential regarding the formation of 3D metal-plastic structures due to the limited materials that can be utilized. To address this issue, researchers have focused on combining 3DP with metallization techniques to create conductive structures. Several studies have attempted to replace 3D metal-plastic structures used in areas such as electronics with various conductive materials, including metallic nanoparticles [20-21], graphene [22-23], multi-walled carbon nanotubes [24] and carbon black [25]. These substances are often used as conductive fillers to obtain modified materials that can be applied to 3DP. Although several types of micro-structured circuits have been successfully fabricated using this approach [26-27], these fillers are typically expensive, and the modified materials tend to exhibit reduced printing accuracy and performance after molding. In addition, the complex preparation process associated with this technology tends to result in slow method development. In contrast, most research has examined the metallization of 3D printed parts via electroplating (EP) [28-29], electroless plating (ELP) [30-31], vacuum evaporation (VE) [32] and sputtering [33]. In this manner, complex 3D printed structures having surfaces with metallic properties have been obtained. Among these methods, ELP is the most cost-effective technique for the deposition of metal onto non-metallic structures. ELP is based on simple wet chemical processing and allows the deposition of a uniform metal coating onto the surface of a part without applying an external electrical potential. However, although many researchers have fabricated microstructures with specific functions using 3DP and ELP [34-38], most of these parts do not meet the complex topological requirements of 3D metal-plastic parts intended to serve as non-conductive or conductive substrates. For this reason, multi-material 3DP (MM3DP) technology has been proposed as a means of achieving pattern selective ELP. Using MM3DP, materials having special functions can be applied at any location of a part to form interconnected 3D patterns that can be used to activate the ELP process. An activation-sensitization pre-treatment is required prior to performing ELP, and this necessity has been exploited to achieve patterned plating. The most straightforward approach to this process is to coarsen specific regions on the material surface by creating microstructures [39-40]. This technique takes advantage of the change in the capacity of the roughened surfaces to adsorb activating/sensitizing Pd catalyst which induces subsequent ELP reaction. However, although selective metallization can be achieved in this manner, the resulting accuracy and resolution are still not satisfactory. Consequently, some researchers have prepared materials with electrode properties that allow more accurate adsorption of activated $Pd^{2+}$ [41] and $Ag^{+}$ [42] seeds, based on generating structures having opposite charges. This method improves the plating accuracy to some extent, although the electrode regions may react with ions in the processing solution when ELP is performed. In contrast to these indirect activation routes, other methods directly add activation seeds comprising ions such as $Pd^{2+}$ [43] and $Ag^{+}$ [44] to the material to fabricate metal-plastic composite structures. Unfortunately, problems related to activator dispersion in the material restrict the application of this technique to just a few 3DP processes. In addition, the quality of the resulting coatings is poor, as is the adhesion of plated mental to the substrate material. Thus, there are remaining challenges related to achieving the fabrication of metallized parts having complex structures in conjunction with high accuracy and performance.

The present study demonstrates a simple means of fabricating metal-plastic functional devices with complex shapes based on a newly developed multi-material digital light processing 3DP (MM-DLP3DP) process. This technique allows the fabrication of microstructures containing active precursors and substrate materials having specific 3D spatial distributions that can selectively activate ELP. The active precursors were prepared by adding a saturated aqueous solution of $Pd^{2+}$ ions to water-washable light-cured resins (either rigid or flexible). Subsequently, complex microstructures with specific topological distributions of material properties (comprising combinations of standard resin with the active precursor) were fabricated using our newly developed MM-DLP3DP apparatus with multiple workstations. After a simple cleaning process, the materials were directly plated and metallized with 3D patterns using ELP. $Pd^{2+}$ ions were homogeneously dispersed throughout the active precursor, and irradiation of the treated resin with patterned UV light caused a photo-initiator to generate free radicals. These radicals initiated a double-bond cross-linking reaction between the monomer and the low-molecular-weight polymer that produced a rigid cured structure in which the $Pd^{2+}$ ions were embedded. In the plating bath, the $Pd^{2+}$ ions exposed at the surface of the material were reduced by $NaH_2PO_2$ to Pd and served as catalytic nuclei after forming Pd particles, following which they induced the targeted metal deposition. This method produced a conjunction layer where the deposited metal is microscopically embedded in the plastic layer. This technique simplifies the production process and reduces costs while providing the capability to manufacture various complex 3D structures with special metal patterns.

## 2. Experimental

### 2.1 Preparation of active precursors

As noted, the technique described herein enables the indirect fabrication of metal-plastic composite structures via the selective 3D deposition of active precursors that subsequently promote ELP reactions. Precursors exhibiting high catalytic activity and positive stability were obtained by adding activation seeds to standard photocurable resins.

Light-curable resins are usually modified by incorporating nano-powders with specific desired properties as fillers, along with dispersants that ensure the uniform distribution of these fillers [45-46]. However, we have found that the stability of modified resins prepared in this manner is low because the filler tends to eventually settle, such that the material becomes inhomogeneous. In addition, the dispersants required for different fillers can be incompatible and may degrade the physical properties of the original substrate materials. To mitigate these issues, we took advantage of the ability of certain water-washable light-cured resins to form homogeneous mixtures with low concentrations (<10%) of aqueous solutions. Thus, the addition of an aqueous solution containing active seeds directly to the resin provided a well-mixed and stable active precursor after sufficient agitation.

$Pd^{2+}$ ions provide superior catalytic activity during ELP and are often used as activation seeds [47-48]. We based this method on the solubility of $Pd^{2+}$ ions in aqueous solutions containing $Cl^-$ ions by employing an activation solution containing $PdCl_2$ powder and $NH_4Cl$. The activated precursors were obtained by homogeneously mixing the activation solution with various resins. This method is both simple and versatile and is applicable to most of the resins in which a polyurethane is the primary component.

Crystalline $NH_4Cl$ was purchased from the FUJIFILM Wako Pure Chemical Corporation while $PdCl_2$ nano-powder (purity: 99.0%) was obtained from the Kanto Chemical Company, Inc. White rigid（Washable, 405 nm, ASIN: JP206000BK510.）, green transparent rigid（405 nm, ASIN: B07CQF6QNM.）, acrylic（405 nm, model: PMMA-like, ASIN: B07SKCNMZX.）, dark gray flexible （405nm, model: SK01F, ASIN: B08T929XVW.） and light gray flexible light-cured resins（405nm, model: SK02F, ASIN: B08T929RXB.）were purchased from the Nova Robot Technology Co., Ltd., Elegoo Co., Ltd., eSun Co., Ltd., Siraya Tech Co. Ltd. and Japan SK Honpo Co., Ltd., respectively. These resins are denoted by resins #1 through 5. Each active precursor was prepared at room temperature (20 °C) by dissolving 3.7 g of $NH_4Cl$ in 10 ml of deionized water, to which 50 mg $PdCl_2$ was added and dissolved with agitation. This produced 10 ml of a saturated activation solution. This solution was allowed to stand for some time, after which a 5 ml portion of the upper, clear part of the solution was removed. Subsequently, a 45 ml quantity of one of the light-cured resins was transferred into a vessel with a magnetic stirrer turning at 1000 rpm, and 5 ml of the activation solution was added dropwise. Following this addition, the mixture was stirred at 1200 rpm for a further 30 min to obtain 50 ml of an active precursor solution.

### 2.2 Fabrication of multi-material nested complex structures via MM-DLP3DP

Compared with direct writing 3DP (DW3DP) [41, 42, 44] and fused deposition modeling 3DP (FDM3DP) [39, 40, 43], DLP3DP allows surface molding with higher resolution, so that parts with smoother surfaces, higher molding accuracy and fill rates can be obtained. Figure 1 provides a diagram summarizing this process, in which the 3DP of a standard resin (serving as the substrate) and of an active precursor forms a part having the desired 3D topology. After cleaning and drying, the surface on which the active precursor is exposed will exhibit catalytic activity. During the subsequent ELP process, the deposition of metal particles from solution will be promoted by $Pd^{2+}$ ions in the active precursor, and selective deposition will occur to form the desired metal pattern.

To enable the manufacturing of multi-material 3D printed parts containing both standard resin and the active precursor, we developed an MM-DLP3DP device incorporating three stations (Fig. 2a). In this apparatus, the printer is able to select from three pools (referred to as material pool A, material pool B and the cleaning pool) and the printing platform can move in both the x and y directions to switch between the different pools. The material pools A, B and cleaning pool hold the standard light-cured resin, the active precursor, and cleaning solution, respectively. The right-hand diagram in Fig. 2a summarizes the structural features of the molding module in the 3D printer. To ensure sufficient molding accuracy and resolution, we utilized the latest generation of 2K black-and-white liquid crystal display (LCD) technology to provide transmissive graphics masking. Ultraviolet (UV) light at 405 nm was passed through a mask to form light fields related to specific slice patterns, thus enabling layered curing and molding of parts. The apparatus comprised an LCD mask capable of moving in the x and y directions together with the z-axis of the 3D printer.

The entire 3D printed component moved as one piece, thus enabling material switching and ensuring that different material topologies in the same part had precise interpositional relationships. Acquiring digital model data for the multi-material part was therefore a critical initial step, even though there are currently no mature multi-material slicing software programs [49] that enable the labeling of material properties in conjunction with specific and complex topologies. In the present work, this issue was addressed by using a modeling-assembly-disassembly-slicing method to assist in the numerical processing of multi-material models (Fig. 2b). In this process, topologies having the same physical properties were created as a single unit and then converted to the standard template library (STL) format. The slicing software assembled units with different properties according to the desired topological relationships, following which these parts were sliced separately to obtain the respective slice data. Because the different parts had defined positional relationships in the part coordinate system, the slice data for the different materials incorporated the required topological relationships. The work reported herein used specially developed control software for slice data processing, the setting of printing parameters and the control of the MM-DLP3DP system. The manufacturing of multi-material parts with 3D active precursors was made possible by performing a series of cycles in which material A was applied, followed by cleaning, followed by the application of material B (refer to Supplementary Material Video 1 for details regarding the manufacturing process).

As shown in Fig. 3, each multi-material part having an arbitrarily complex structure, regardless of its structural characteristics, could be analyzed in terms of the different material topologies. This analysis involved both the interlayer multi-material stacking topology and multi-material nested topology. In the case of the former, the various materials were nestled between layers, thus eliminating the need for the printer to perform a cleaning and material switching process at each slice. However, when processing the latter topology, the printer had to cycle through printing of material A, printing of material B and a cleaning process for each slice. Since all materials used in the part had the topological characteristics described above, the proposed process allowed the fabrication of various complex parts, although it should be noted that the fabrication of certain components required the assistance of support structures.

### 2.3 Selective 3D metallization using ELP

During processing, the printed part was firmly adhered to the printing platform and so, because the platform was made of metal, a small amount of metal residue could remain on the base of the part after it was removed. This residue could promote the precipitation of Ni and so degrade the accuracy of subsequent selective ELP and interrupt the metal distribution pattern. In addition, uncured resin remaining on the surface of the part could obscure $Pd^{2+}$ ions inside the resin and so affect the catalytic activity of the active precursor during plating. For these reasons, it was vital to clean parts after the 3DP. In the present work, the cleaning solution was composed of 40 % ethanol (analysis pure), 50 % acetone (analysis pure) and 10 % dilute sulfuric acid (40 wt %) in volume. The alcohol and acetone in this mixture dissolved any uncured resin on the part surface (both the standard resin substrate and the active precursor) while the sulfuric acid removed residual metal powder adhering to the bottom of the part.

Unlike the ELP processes used in other studies [39-42,44], our method does not require the cleaned finished parts to be roughened or sensitized. Consequently, the cleaned and dried parts can be directly immersed in the plating bath. The Ni plating bath employed in this process had the primary components summarized in Table 1; it had a pH of 9 and was held at 70 °C (see Supplementary Material for details regarding the electroless Cu plating). Within each printed multi-material part, the active precursor (in which $Pd^{2+}$ ions were homogeneously dispersed) was distributed on the resin substrate in a specific 3D topology. Upon immersion of the part in the bath, the exposed $Pd^{2+}$ ions on the surface were initially reduced to Pd monomers that served as catalytically active metal nuclei to initiate the ELP reaction in specific microscopic regions, and so achieve targeted Ni metal deposition. The reactions shown in equations (1)-(4) comprise the mechanism by which the reactive precursor catalyzes the directed deposition of Ni metal. In this process, hypophosphite is oxidized in solution to generate adsorbed hydrogen atoms on the surface of the substrate. Immediately afterwards, these hydrogen atoms reduce Ni ions in the solution. Because hydrogen atoms are adsorbed on the substrate surface, the reduced Ni is naturally deposited on the same surface after 5-10 min to form a coating. In addition, since $Pd^{2+}$ ions are embedded on the surface of the active precursor portions of the part, there is no overflow or deviated deposition of the plating layer due to migration of the catalyst during the ELP process. Because time-consuming pretreatment of the part surfaces is not required, the original surface morphology is maintained and so a more accurate plating pattern can be obtained (refer to Supplementary Material Video 2 for details of the electroless Ni plating process).

**Table 1 Electroless Ni plating bath composition and operating conditions**

| Component | Concentration [mmol/L] |
|---|---|
| $NiSO_4 \cdot 6H_2O$ | 60 |
| $NaH_2PO_2 \cdot H_2O$ | 240 |
| $C_6H_5Na_3O_7 \cdot 2H_2O$ | 200 |
| $H_3BO_3$ | 500 |
| $H_2SO_4$ | pH adjustment |
| NaOH | |
| pH | 9.0 |
| Temperature | 70 °C |

$$H_2PO_2^- + H_2O \rightarrow HPO_3^- + H^+ + 2H \quad (1)$$

$$H_2PO_2^- + H^+ + 2H \rightarrow 2H_2O + P \quad (2)$$

$$2H \rightarrow H_2 \uparrow \quad (3)$$

$$Ni^{2+} + 2H \rightarrow Ni + 2H^+ \quad (4)$$

## 3. Results and discussion

### 3.1 Properties of reactive precursors combined with commercial photocured resins

After modification, resin #1 changed from its original white color to a light yellow. The uniform distribution of this coloration indicated that $Pd^{2+}$ ions were homogeneously dispersed throughout the material. Similarly, the #5 resin changed from gray to yellow gray. After three days of standing, neither material exhibited precipitation, indicating the high stability of these precursors. Figure 4a shows the UV absorption spectra obtained from the #1 resin before and after modification. Both spectra are basically equivalent with peak absorbance at approximately 405 nm. These results demonstrate that the addition of the $Pd^{2+}$ solution did not change the basic properties of the resin, especially the molding properties. It is also apparent that the addition of the $Pd^{2+}$ solution did not decrease the light sensitivity of the original resin. Similar results were obtained from the analysis of the #4 resin before and after it was modified, as shown in Figure 4b.

### 3.2 Complex 3D metal-plastic composite structures

Circuit boards are traditionally made from flat modules or a combination of multiple flat modules such that the processing surfaces are 2D planes or combinations of such planes [50]. These structures are relatively simple and so the manufacturing process is not complex but has limited applications. The demands for more structurally complex parts with regular cylindrical or freeform processing surfaces requires improved manufacturing capabilities. To illustrate the fabrication capabilities of our new approach, metal-plastic composite parts having representative structures were fabricated and then processed via ELP. Compared with multi-material 3DP using multiple nozzles, the technique demonstrated herein provides higher resolution and thus allows the construction of micro-structured surfaces with special functions. This is important because metallized micro-structured surfaces have numerous potential applications.

To demonstrate this, we constructed three different surface structures on circular substrates. These comprised a circular microstructure with a radius of 200 μm, an ortho-hexagonal microstructure with an inner circle radius of 200 μm and a circumferential groove microstructure with a groove width of 400 μm (Fig. 5a). After printing and cleaning, these microstructures could be clearly seen under a high magnification microscope (Enlarged view in the upper right corner of the picture). After plating, metallic Ni uniformly covered the microstructure surface and a 3D profile consistent with the active precursor was obtained. Figure 5b shows a metal skeleton ball structure with a circular base fabricated using this process. This microfine structure is interconnected in an irregular manner and, after plating, the skeleton was covered with Ni metal. From the partial enlargement shown on the right-hand side of this image, it is obvious that the structure was evenly plated inside and outside and had a metallic luster with no under-plating or missing plating. The left-hand side image in Fig. 5c shows a 3D printed Eiffel Tower model with the tip and middle part made of the active precursor material (shown in light yellow) and the rest made of standard resin (shown in green). After plating, Ni metal was precisely deposited on both the tip and central surface of the tower. The partial enlargement shows the precise structure of the metal part, which evenly covered the top and middle of the tower. It is also evident that the original hollow microstructure remained intact and that the interior and exterior of the hollow structure in the middle of the tower were covered with Ni metal having a metallic luster. The manufacturing of these multi-material nested parts demonstrates the capabilities of this technology.

Figure 5d shows a multi-material five-sided part with a specific metal topology distributed on four faces. This structure is highly complex, with the metal portions and the resin substrate nested within one another such that there is both interlayer and intralayer nesting. Despite this complexity, clear boundaries are evident between these portions of the part, confirming that precise selective metallization was achieved. The right-hand side of the figure presents an optical microscopy image of the Ni metal-substrate junction that demonstrates a uniform plating distribution and no spillage or contamination of the Ni plating. Figure 5e presents an image of a domed structure with six curved metal bands evenly distributed around its surface. Figure 5f shows a U-tube part having an internal hexagonal mesh made of Ni. This internal mesh structure is particularly difficult to fabricate by most other manufacturing technologies. This highly complex part could not be fabricated by conventional methods using laser etching or 3DP but was obtainable based on slicing and material switching that permitted deposition of the active precursor inside the U-tube (left-hand image). During the ELP process, the plating solution was able to flow into the tubular part to form the mesh in the cavity via the deposition of Ni (middle image). The right-hand side of this figure indicates the intricate structure of the metal mesh after removal from the part. The mesh had a thickness of 1 mm and contained hexagonal holes each with an inner diameter of 500 μm. Furthermore, we made a double-layer 3D hollow part (Fig. 5g) in which the large ball (radius 15 mm) has a triangular hollow hole wrapped around the small ball (radius 8 mm) with a regular hexagon hollow hole. In this part, the plating solution enters through the holes of the outer large ball, so that the selective Ni deposition of the inner small ball can be successfully achieved. The enlarged image (Right side of the figure) shows the metallic luster of the inner ball and illustrates a good Ni coating. These two parts demonstrate the possibility of manufacturing items with complex internal metal structures, thus broadening the potential applications of this technology.

This new technology was also used to manufacture flexible 3D electronics as a proof of concept. The present surface molding 3DP process maintains the manufacturing success rate of flexible parts, as no support is required to obtain the desired flexible structure. Figure 5h demonstrates a flexible carbon nanotube structure, the central part of which was selectively applied using ELP. The high-resolution image here shows the junction between the metal and resin in this item. The metal plating was evidently uniformly distributed and the boundaries between the different materials are well defined, indicating the effective formation of a flexible metal-plastic composite structure using this method. After being stressed, the Ni metal plating remained intact and did not break. Figure 5i shows a flexible wearable hoop-shaped structure. The enlargement demonstrates the precise deposition of the metal plating on the surface of the knotted part. This item was found to be highly flexible, and the metal structure did not break in response to bending deformation. In addition to the above parts, we designed and fabricated various 3D stereoscopic circuits to replace conventional PCB circuits, thus illustrating the potential of the proposed process for application in the field of 3D micro-nano electronics.

Figure 6a presents diagrams and a photographic image of a light-emitting diode (LED) circuit with an irregular 3D structure. The circuit board model in the figure shows that the substrate had an irregular surface profile while the wires had a complex 3D alignment. After printing, soldering, and powering up at 3.3 V, the LED was found to function with a normal level of brightness. This result confirmed that the metal wire structure distributed on the substrate was suitably conductive. Figure 6b shows a double-layer 3D circuit with a through-hole structure in which the width of the conductor was 800 μm. The diameter of the through hole was 500 μm and the inner wall was covered with Cu metal that connected the inner and outer layers of the circuit. This circuit had a complicated 3D wire distribution that reduced the size of the device compared with a standard circuit and increased the electron transmission efficiency. These improvements would be expected to increase the extent of integration in 3D electronic devices and so to have numerous practical applications. Figure 6c provides plots summarizing the impedance characteristics of Cu and Ni plating obtained using the proposed process and confirms that the resulting 3D metal wires exhibited electrical conductivity that meets the requirements for use in electronic devices.

Sample parts with very small components were also designed and manufactured to examine the accuracy and resolution achievable via selective metallization. Figure 7a shows images of Ni wires having widths of 1 mm, 500 μm, 100 μm, 50 μm, and 40 μm. High-magnification microscopy showed that the Ni was precisely distributed on these active precursor wires, thus establishing the exceptional resolution of the printing method and the effectiveness of selective metal deposition. Cu plating on a circuit board is shown in Fig. 7b. From this image, it is apparent that the proposed method provides a manufacturing resolution of at least 40 μm, which is sufficient to meet the processing requirements of most electronic products.

### 3.3 Microscopic characterization of 3D selective metal topologies

Microscopic characterization of the metallic plating generated by the present metallization method was considered vital. The effectiveness of ELP induced by active precursors and the quality of the resulting plating was assessed by fabricating samples having dimensions of 20 mm (length) × 10 mm (width) × 5 mm (height). These specimens had planar, circular, ortho-hexagonal, and annular grooved micro-nano structures with metallic coatings.

Figures 8a-c present microscopic characterization results for a planar sample after a 5 min plating process. The sample was uniformly covered with Ni and a metallic luster can be observed in the low-magnification light microscope image (right side of the figure 8a). SEM images acquired with a 100 μm scale bar show the surface profile of the Ni layer, which is flat in most places, in agreement with the planar profile of the active precursor. Some minor protrusions are evident due to the accumulation of Ni metal caused by high local $Pd^{2+}$ concentrations. Higher magnification shows that the Ni metal particles were closely packed together with a more even distribution of crystals and less discontinuity. Figures 8d-f present images of the specimen obtained after a 6 min plating of a circular surface having a radius of 200 μm and a depth of 200 μm. Figures 8g-i show the plating on the surface of an annular groove with a width of 200 μm and a depth of 200 μm while Figs. 8j-l show images of a plated hexagonal microstructure having an inner radius of 200 μm and a depth of 200 μm. All three-

parts exhibit complete and uniform plating with a metallic luster. The SEM images with a 100 μm scale bar demonstrate that the Ni metal particles all grew along the surface profile in a uniform manner. The plating did not make the original surface structure unclear due to uneven thickness, again demonstrating the uniformity of the plating. SEM images with the 500 nm scale bar clearly demonstrate the morphology and distribution of the Ni metal particles. The distribution of Ni crystals on the surfaces of these microstructure samples was more disordered than on the planar specimens but was still compact. In the active precursor regions, $Pd^{2+}$ ions were embedded in the cured resin, but the bonding interactions between the molecules in flexible resin were not as strong in these resins compared with the rigid resins after curing. Therefore, the $Pd^{2+}$ ions in the flexible active precursors precipitated more easily during plating such that shorter plating times were required. Figures 8m-o show the results for 3 min Ni plating of a flexible active precursor specimen. The resulting metalized microstructure not only increased the functionality of the part but also strengthened the bond between the plating and the substrate. Tape tests showed that Ni metal on microstructure surfaces exhibited a higher bonding force compared with metal on flat surfaces, as a consequence of the higher roughness associated with the microstructures. Compared with the rigid samples, the flexible samples showed more consistent plating with higher integrity, and with 100 μm scale bar SEM observations indicated that the latter had much rougher surfaces. This phenomenon was attributed to corrosion of the flexible surface by the acidic component of the plating solution. Corrosion is also evident in the 500 nm scale bar SEM images, which show a significant increase in the number of Ni metal particles in the same area, with continuous and tight clusters of crystals. This occurred because more $Pd^{2+}$ ions were exposed when the sample was inadvertently roughened.

We also found that this method, in which catalytic seeds were incorporated into standard resins, allowed us to obtain metal layers with higher confinement and bonding strength. Ni-plated samples were processed using a focused ion beam to obtain cross sections allowing better examination of the microstructures at the metal-resin junctions. Analyses by transmission electron microscopy (TEM) and energy-dispersive X-ray spectroscopy (EDS) were subsequently performed.

Figure 9a presents a TEM image acquired at the boundary between the resin and the Ni metal layer. After plating, the cross section shows the Ni metal, conjunction, and resin layers. Figure 9b indicates the distribution of carbon along the cross section containing the resin boundary. Carbon was more abundant in the resin layer and significantly less abundant in the intermediate layer, such that the partition between these two phases is highly consistent with that in Fig. 9a. This result supports the existence of a conjunction layer. During the ELP process, the sodium hypophosphite in the plating solution reduced $Pd^{2+}$ ions to Pd particles that were attached to the surface. The Pd in the interlayer catalyzed the deposition of Ni such that the Ni metal migrated into the resin to generate an interlayer of Ni embedded in the material. This effect improved the adhesion of the Ni plating.

### 3.4 3D printed metal-plastic composite parts for electronic devices

The proposed technology can combine non-metallic materials having different functions with specific metal patterns to generate metal-plastic composite parts. Based on this technology, we propose a series of potential applications.

### 3.4.1 3D printed strain gauge

A metal-based resistance strain gauge works on the principle of the resistance strain effect. Specifically, when a Cu wire is subjected to stress, its resistance will change in proportion to the degree of stretching or compression [51,52,53]. In conventional strain gauges, a resistive measurement grid is usually laminated between a carrier and cover film, and this foil strain gauge is bonded using an adhesive to the component being assessed. Strain in the component is transferred to the measurement grid via two intermediate layers. In principle, the measurement grid should be placed as close as possible to the component surface in order to avoid force transmission losses. In such devices, variations in bonding thickness will be reflected in the degree of force transmission, meaning that the response of the strain gauge may change. The present technique allows strain gauges to be printed inside complex parts. This not only ensures consistent mechanical deformation between the strain gauge and the measured object, but also enables the design and manufacture of specific strain gauges that match the actual 3D shape of the object. Figure 10a shows a strain gauge designed by our group that is integrated with the measured object. The overall part is a cuboid with openings, and the strain gauge is in the interior of the base. Electrical signals generated when the internal strain gauge is deformed are extracted via a hole in the device. Figure 10b presents images indicating the structure of the internal strain gauge. Using a 3D printing and Cu plating process, we were able to form metal electrodes having a pitch and width of 500 μm. When the strain gauge was bent and deformed as the object being tested was subjected to force, the overall length of the through-electrode changed and so its resistance varied. By measuring the change in voltage, we could calculate the strain and thus the amount of deformation. Figure 10c provides a photographic image of the strain gauge after it has been deformed. Because we used an acrylic-type UV resin, the strain gauge was highly flexible and so did not break when bent. Figure 10d plots the voltage generated by the strain gauge as a function of the load, as obtained using a star-shaped circuit. Figures 10e and 10f summarize the calculated strain and deformation values, respectively. When the load is 10 g, the strain value of the sensor obtained through the experiment is 0.003. This data is basically consistent with the result of the simulation calculation (the upper left corner of Fig. 10 e)), which verifies the effectiveness of the sensor. This device could therefore be used to perform strain measurements when correcting for variations in temperature and material properties.

### 3.4.2 3D printed piezoelectric sensor

The proposed technology allows the construction of metal wires on the surfaces of various functional materials. In the present work, the polymer PVDF-TrFE [54-56] was added to an elastic light-cured resin to produce a material with piezoelectric properties that could also be used as the active precursor. Figure 11a presents diagrams showing the functioning of a 3D printed piezoelectric sensor consisting of a thin film substrate made of a piezoelectric material and comb electrodes. When this type of sensor is subjected to an external force along its sensitive axis, electric charges of opposite polarity are generated on the two adjacent electrodes, corresponding to an electric charge source (that is, an electrostatic generator). The piezoelectric effect allows the device to sense deformation and thus can be used to measure deformation, strain, angular changes, and other variables. Figure 11b provides photographic images showing the device being bent at various angles, while the resulting voltage values are presented in the oscilloscope trace in Fig. 11c. Figure 11d shows a plot indicating the piezoelectric effect for the sensor in the stretched state. As the amount of deformation was increased (that is, the length of the central axis of the upper surface increased), the voltage generated via the piezoelectric effect also increased. This voltage was directly correlated with specific variables and could provide accurate measurements with the support of additional filtering and amplifier circuits. Using this principle, we could therefore achieve accurate measurement of certain physical quantities. Compared with conventional piezoelectric sensors, piezoelectric sensors made by 3DP require a less complex manufacturing process that can be integrated with that for the object being examined to permit a wider range of applications.

### 3.4.3 Wearable 3D printed flexible ECG electrodes

An electrocardiogram (ECG) provides information regarding the electrical activity of the heart and is widely used for the diagnosis and analysis of many diseases [57-61]. Typically, a complete ECG measurement system is primarily made of electrodes, digital processing circuits and data analysis terminals. Among these, the electrodes are the key components that affect the quality of the electrical signal received from the heart. We designed a wearable flexible 3D printed ECG electrode and fabricated this device using the new technology presented herein. Figure 12a shows the working principle of this device, which used a dual-lead measurement method in which the host ECG and vice ECG were worn on the right and left wrists of the subject, respectively. The electrodes acquired the ECG signal and sent it to a computer via a Bluetooth module configured on the host ECG, and the computer in turn processed and displayed the signal.

Figure 12b provides images of the 3D printed strap-like wearable device having a specially patterned metal electrode, a circuit for connecting with a portable battery, and the Bluetooth module. This configuration eliminated measurement errors and inconvenience caused by wires and made it easier to measure ECG signals while the subject was in motion. Figure 12c shows plots of five different ECG signals obtained with the proposed ECG electrode with the subject in the resting state. The five plots essentially overlap and that the waveforms are those expected from the human heart. Figure 12d shows the measurement signals obtained with the ECG electrode during writing, swinging of one arm and clicking of a computer mouse. Compared with a conventional ECG, the proposed electrode offers an easier measurement process together with efficient output of the resulting data.

## 4. Conclusions

This work demonstrated a new approach to constructing precise metal patterns on the surfaces or interiors of 3D plastic parts with arbitrary complex shapes, along with potential applications. By modifying standard resins, we prepared active precursors that were able to promote the ELP process. An MM-DLP3DP apparatus was developed to permit the fabrication of micro-nano 3D metal-plastic component structures, and the proposed technology was employed to fabricate various parts as a demonstration of manufacturing capability. These parts were primarily multi-material and involved nesting layers, including microporous and small hollow structures with the smallest size being 40 μm. A 3D circuit with a double-sided structure connected by a through hole was manufactured to illustrate the manufacturing capabilities of the proposed technology. Finally, a series of sensors operating on different principles (a 3D printed strain gauge, 3D printed piezoelectric sensor and wearable 3D printed flexible ECG electrodes) was fabricated to illustrate the superiority of this technique. Compared with conventional processes, this new technology allows integrated manufacturing of the sensor and the object to be measured, thus avoiding measurement errors and complex processes caused by assembly. As more specialized light-cured resins (such as acrylonitrile-butadiene-styrene, acrylic and silicon materials) become available, it should be possible to fabricate 3D structures with unique properties. In addition, a variety of metals (including Ni, Co, Cu, Au, Ag and Pt) could potentially be deposited in targeted patterns by means of active precursor-induced plating. This process enables the 3D nesting of a variety of composites and thus has promising applications, especially in MEMS, sensors and robotics, wearable devices, and 3D precision electronics.

**Acknowledgments:**

The authors thank YUITOO Researcher and Research Organization for Nano & Life Innovation, Waseda University for supporting to experiments. Thank ENOMOTO Researcher and Kagami Memorial Research Institute for Materials Science and Technology, Waseda University for helping to experiments.

**Funding:**

This work was supported by Singapore Ministry of Education [MOE2017-T2-2-067], JST-Mirai Program Grant Number JPMJMI21I1, and Kakenhi Grant Number 19H02117 and 20K20986, Japan.

**Credit authorship contribution statement:**

Kewei Song: Methodology, Investigation, Writing - original draft, Writing - review & editing, Visualization, Data curation. Yue Cui: Writing - review & editing, Visualization, Data curation. Tiannan Tao: Methodology, Investigation, Data curation. Xiangyi Meng: Investigation, Review & editing. Michinari Sone: Methodology, Investigation. Masahiro Yoshino: Methodology, Investigation. Shinjiro Umezu: Conceptualization, Methodology, Investigation, Writing - review & editing, Visualization, Supervision, Project administration, Funding acquisition. Hirotaka Sato: Conceptualization, Methodology, Investigation, Writing - review & editing, Visualization, Supervision, Project administration, Funding acquisition.

## Competing Interest:

The authors report no declarations of interest.

## Figures:

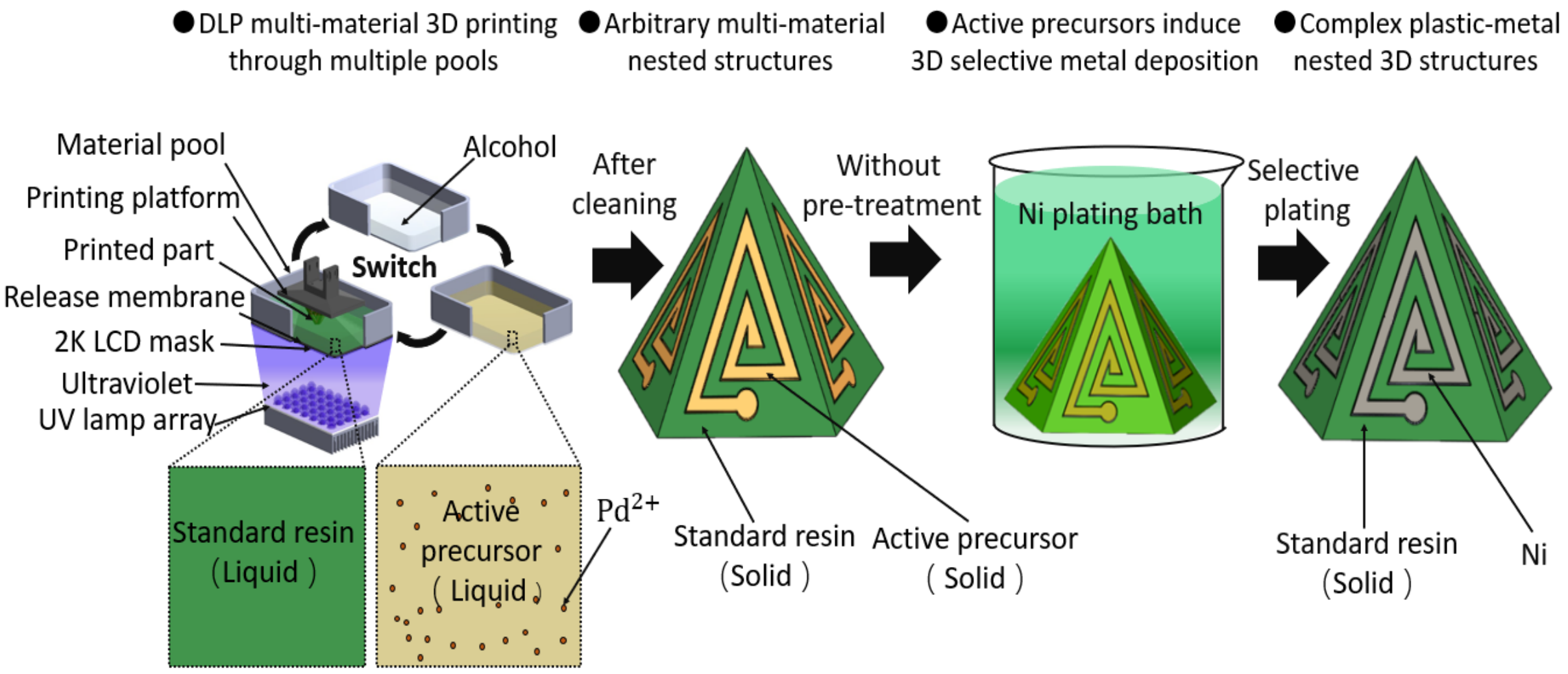

**Fig. 1. The use of MM-DLP3DP and ELP to manufacture 3D metal-plastic composite functional devices having complex shapes.** 3D printed standard resin and active precursor parts can induce ELP reactions to achieve selective metal deposition.

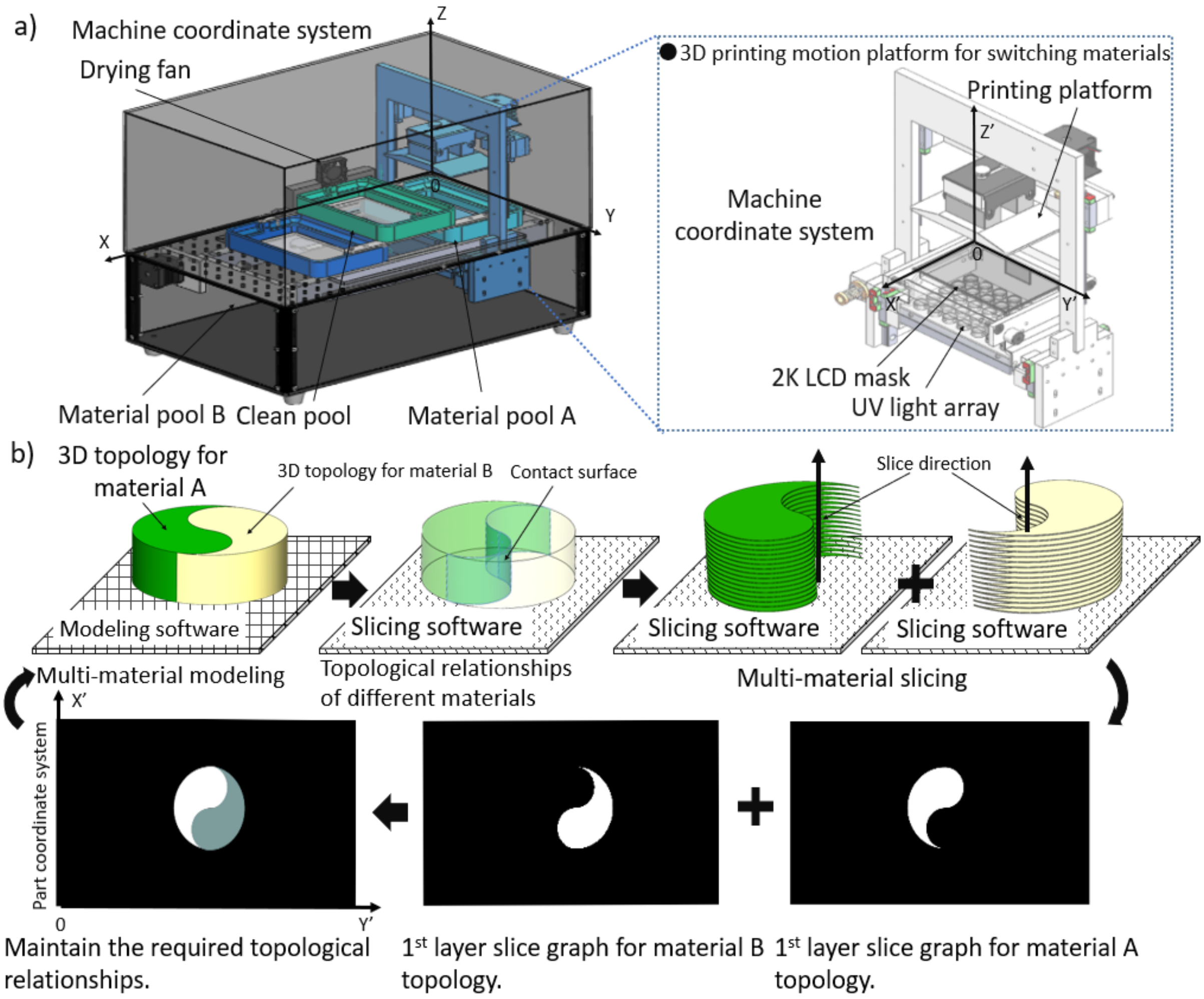

**Fig. 2. Diagrams of the MM-DLP3DP apparatus and process flow.** a. The structure of the MM-DLP3DP apparatus incorporating three pools that enable the manufacture of multi-material parts. b. Numerical processing of multi-material models in which the slicing data are obtained by treating different material topologies in the same part as units with separate slicing.

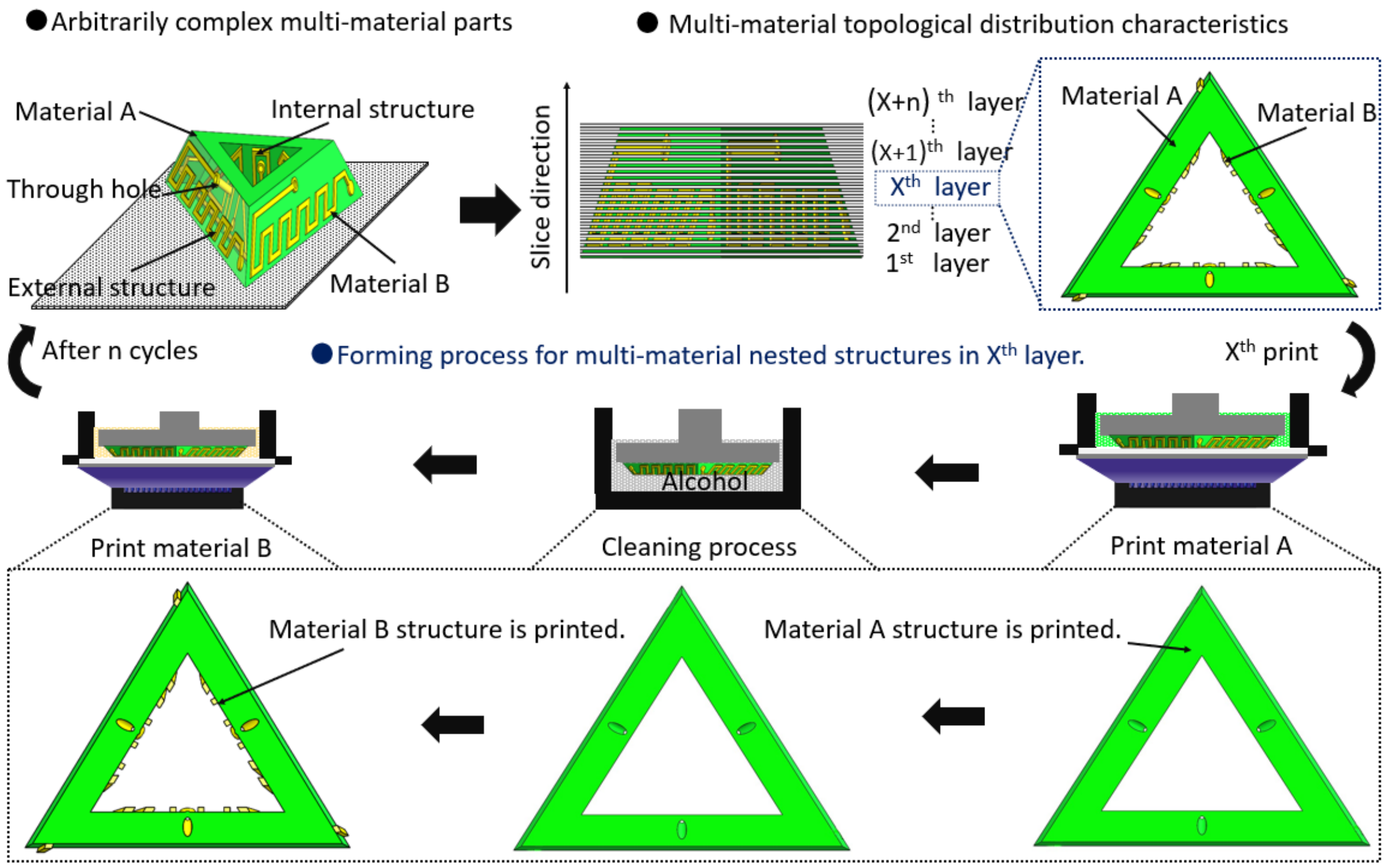

**Fig. 3. Structural characteristics of an arbitrary complex multi-material model and the printing process.** The multi-material model of an arbitrary complex structure can be divided into interlayer multi-material stacking and multi-material nested topologies according to the desired distributions of the material topologies.

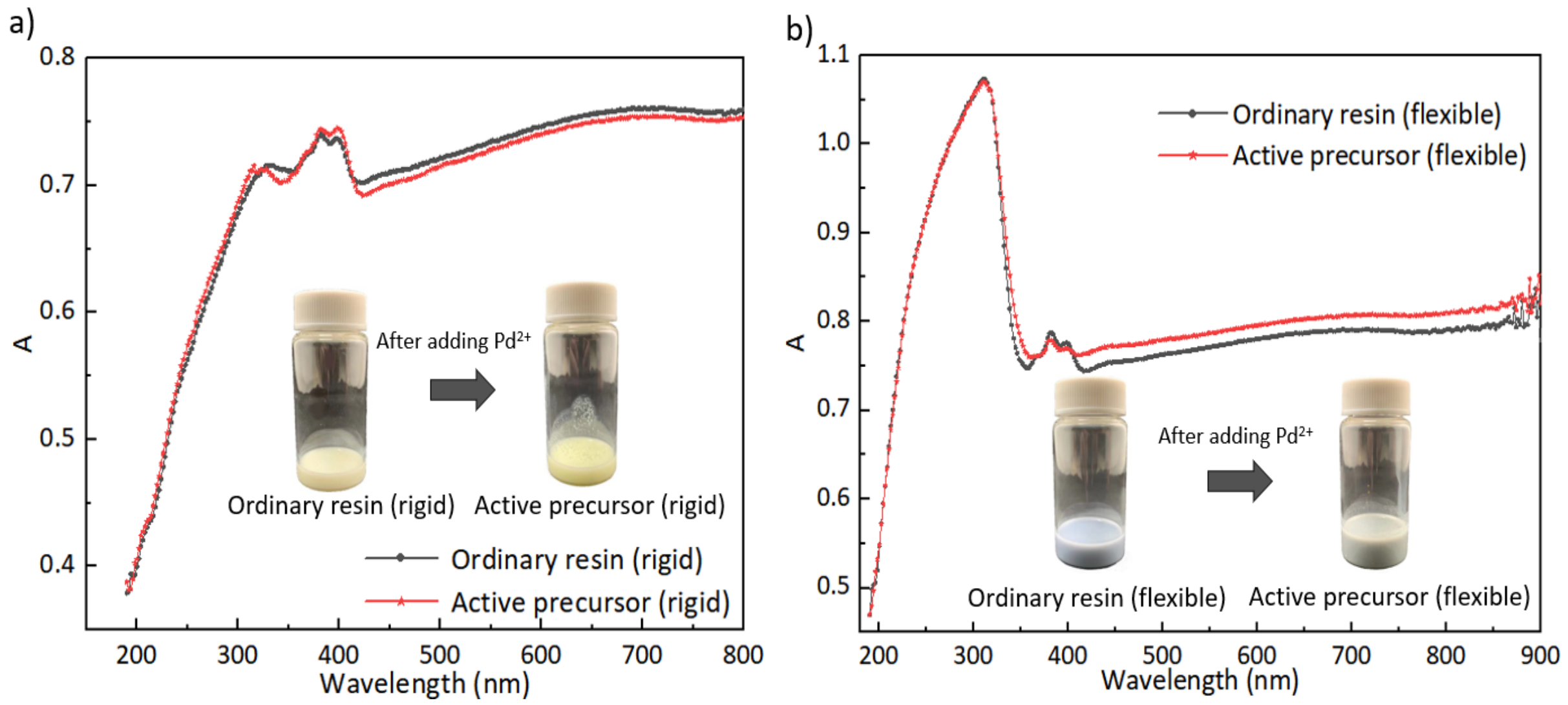

**Fig. 4. UV absorption spectra of resins and active precursors.** Photographic images and spectra of a. rigid and b. flexible light-cured resins before and after modification with $Pd^{2+}$ ions, demonstrating minimal change in both cases.

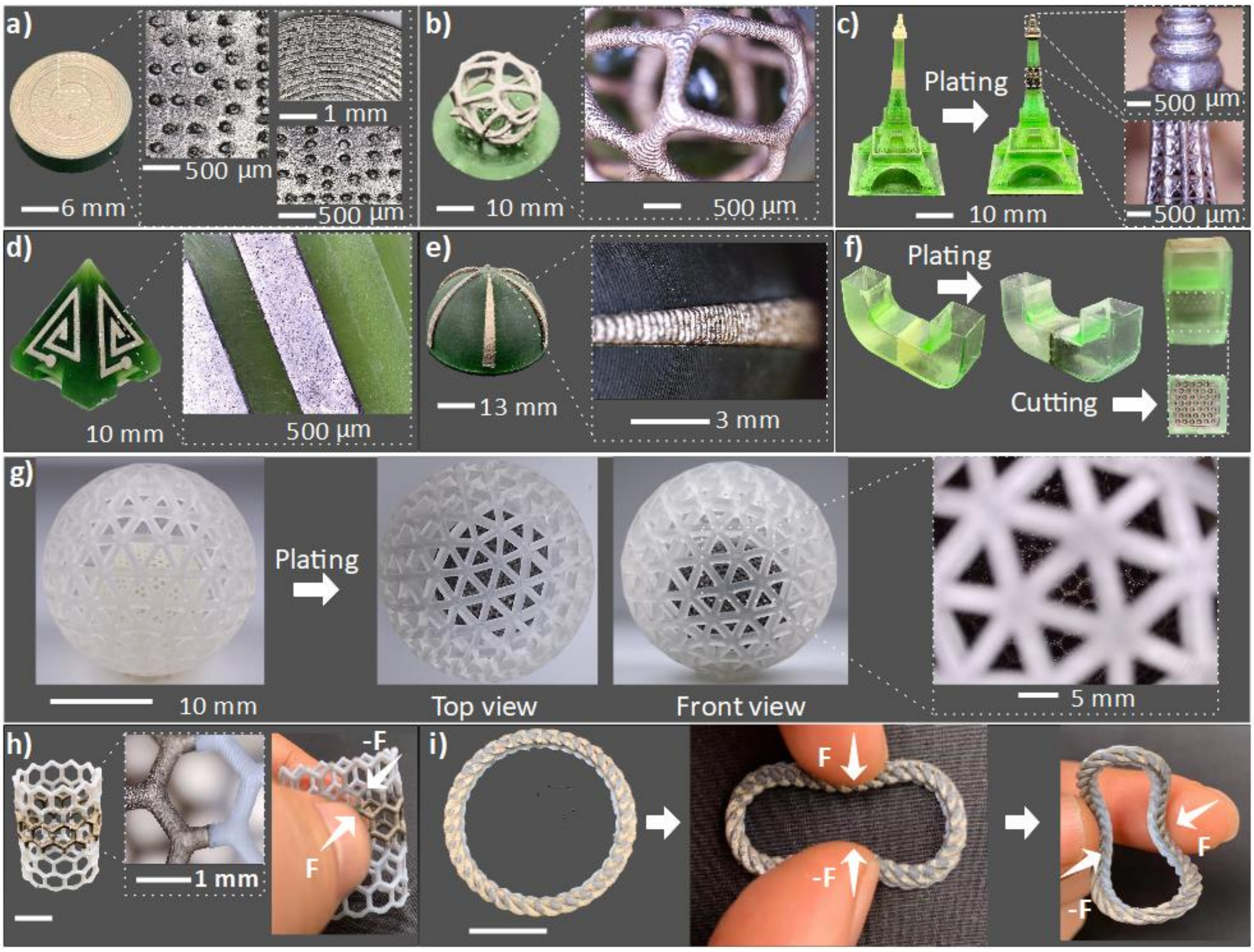

**Fig. 5. Examples of 3D metal-plastic composite structural parts with complex structures manufactured using this technology.** a. Three micro-structured metallized surfaces each with a thickness of 200 µm: circular with a radius of 200 µm, ortho-hexagonal with an inner circle radius of 200 µm, and circular groove with a groove width of 400 µm. b. A skeletonized ball structure with a circular base that has been covered with Ni metal. c. An Eiffel Tower with a microfine structure in which selected parts have been uniformly covered with Ni plating. d. A five-sided structure with a complex Ni metal distribution and multi-material nesting between layers and within layers. e. A dome structure with an arc-shaped metal distribution. f. A U-shaped tube with an internal metal mesh having a thickness of 1 mm and square hexagons each with a 500 µm inner diameter, demonstrating the construction and metallization of a small structure within the resin. g. A double-layer 3D hollow nesting part, a large ball (The radius is 15 mm.) with a triangular hollow hole wraps a small ball (The radius is 8 mm.) with a regular hexagon hollow hole. Among them, the small balls inside are selectively deposited with Ni. h. A flexible carbon nanotube structure with selective 3D metallization, showing a lack of fracture after deformation. i. A wearable structure with a complex shape and Ni plating that remains tightly bonded to the flexible substrate after deformation. These micro-sized hollow or complex heterogeneous structures as small as 400 µm confirm the precision manufacturing that is possible utilizing this method.

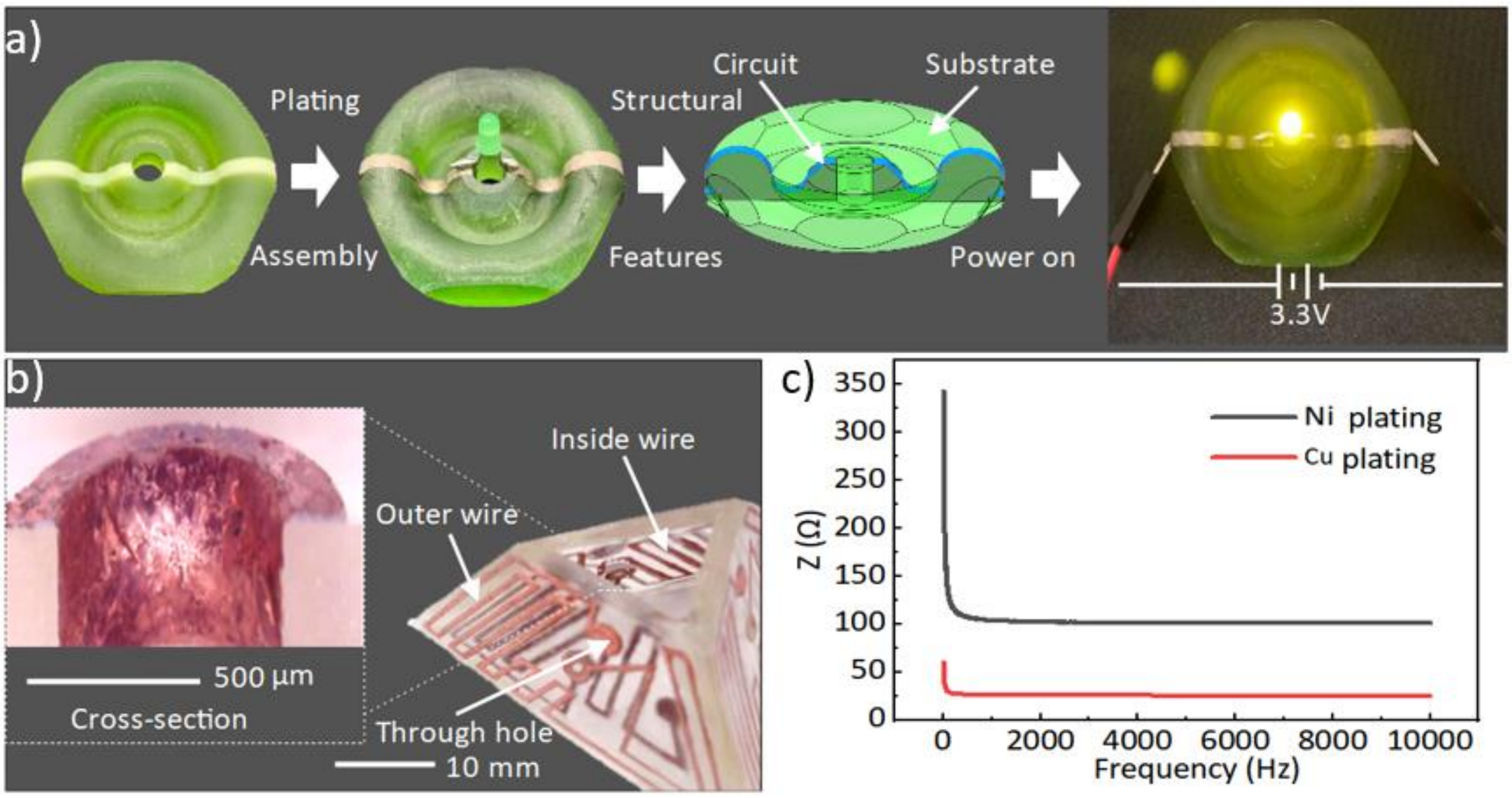

**Fig. 6. 3D circuit board with a specific metal wire topology.** a. An LED stereo circuit with a complex Ni metal wire topology. This LED emitted light normally after power was applied. b. A double-layer 3D circuit structure with a complex 3D copper wire structure which cannot be fabricated by traditional process. c. The impedance characteristics of Ni and Cu coatings generated using the proposed process.

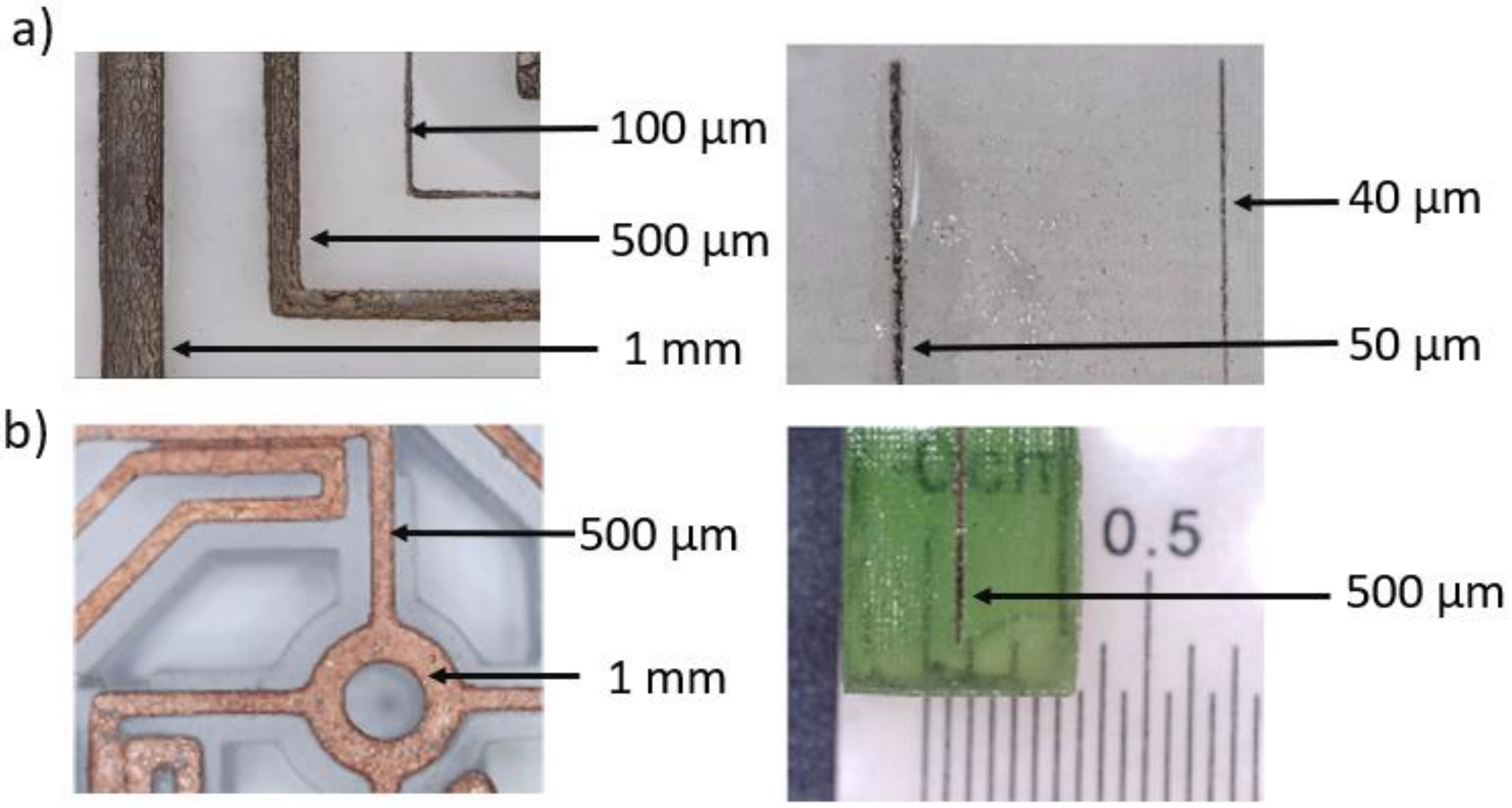

**Fig. 7. Resolution testing of metal-plastic composite structures.** a. Ni-plated structures with widths of 1 mm, 500 µm and 100 µm and with widths of 50 and 40 µm. b. A circuit board part with a width of 1 mm incorporating 500 µm copper wires.

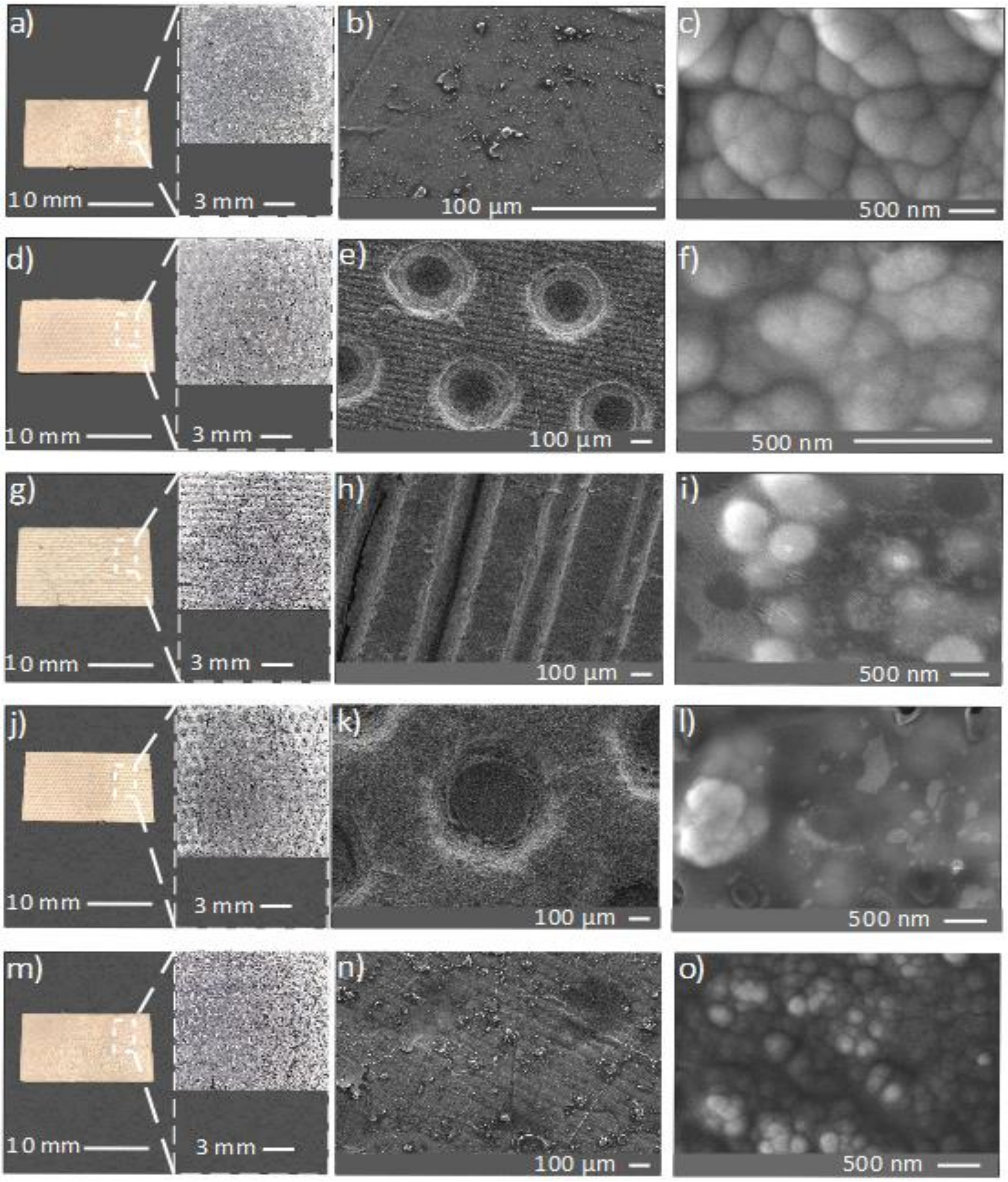

**Fig. 8. Microscopic characterization of Ni plating on active precursor surfaces.** a. Standard flat Ni plating. b and c. SEM images of the indicated area with different magnifications. d. Ni plating on the surface of a circular microstructure. e and f. SEM images of the indicated area with different magnifications. g. Ni plating on the surface of an annular groove microstructure. h and i. SEM images of the indicated area with different magnifications. j. Ni plating on the surface of an ortho-hexagonal microstructure. k and l. SEM images of the indicated area with different magnifications. m. A flexible sample with Ni plating. n and o. SEM images of the indicated area with different magnifications.

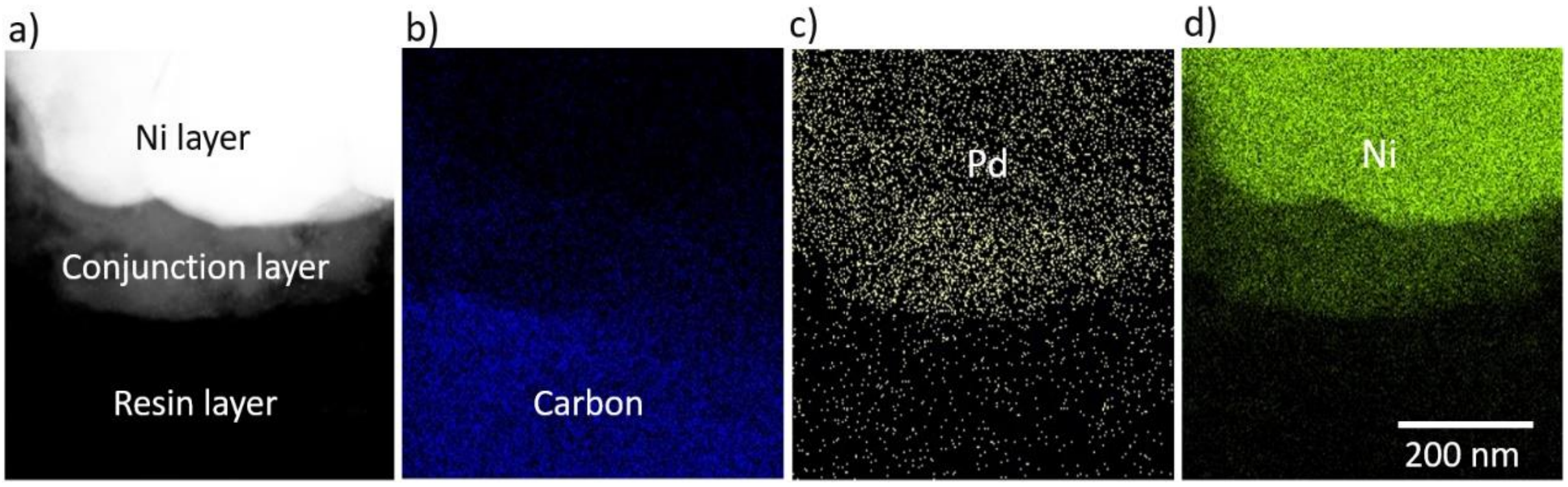

**Fig. 9. Elemental analysis of resin-metal cross section surfaces by EDS.** a. The microstructure of the cross section showing resin, conjunction, and Ni layers. The distributions of b. carbon, c. palladium and d. nickel.

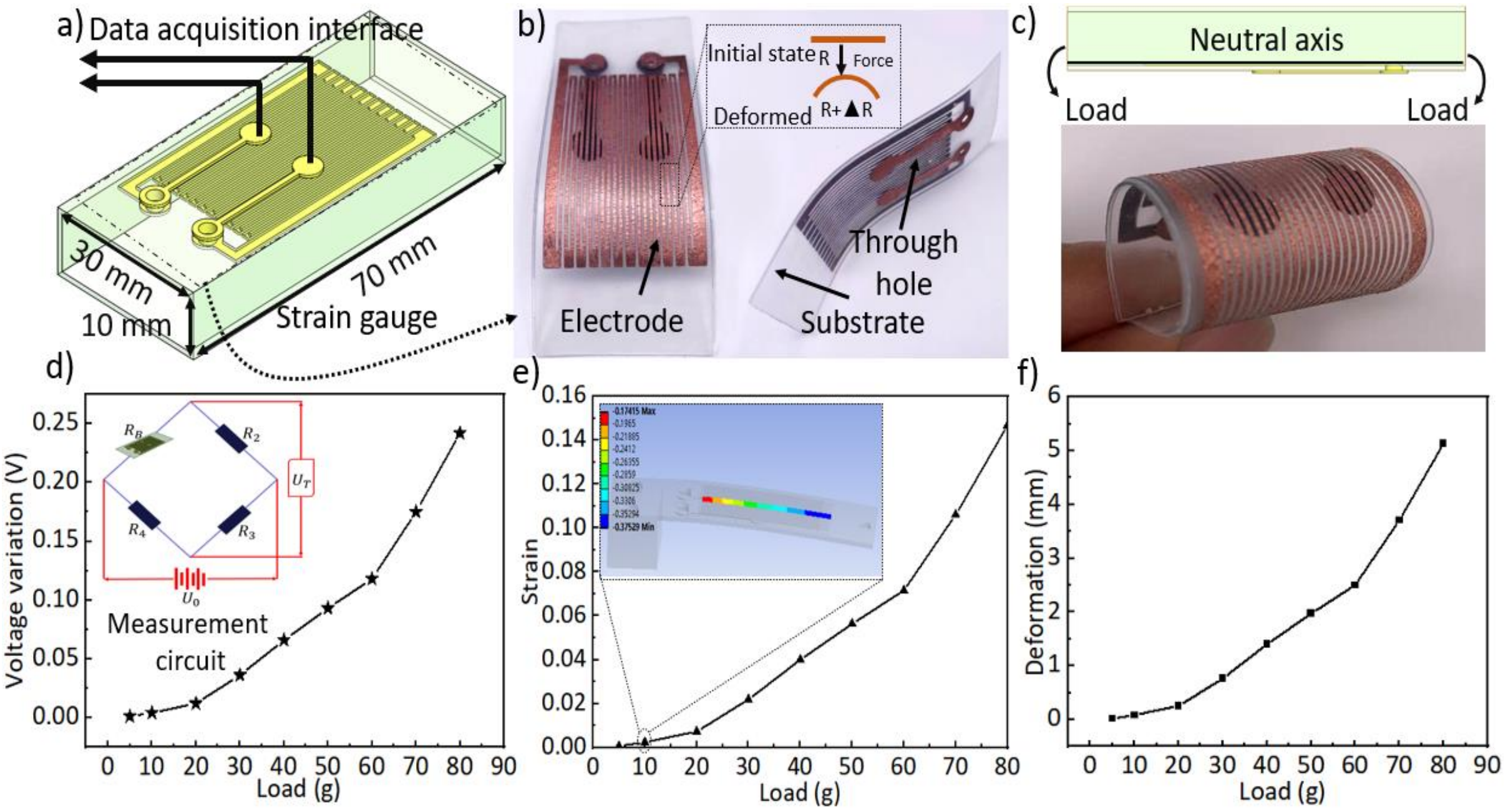

**Fig. 10. A 3D printed integrated strain gauge.** a. Structures of the 3D printing integrated strain gauge and the measured object. b. The structure and measurement principle of the gauge. c. Bending deformation of the gauge under stress. d. Voltage as a function of load for the gauge. e. Strain values calculated from experimental data. f. Experimental deformation data.

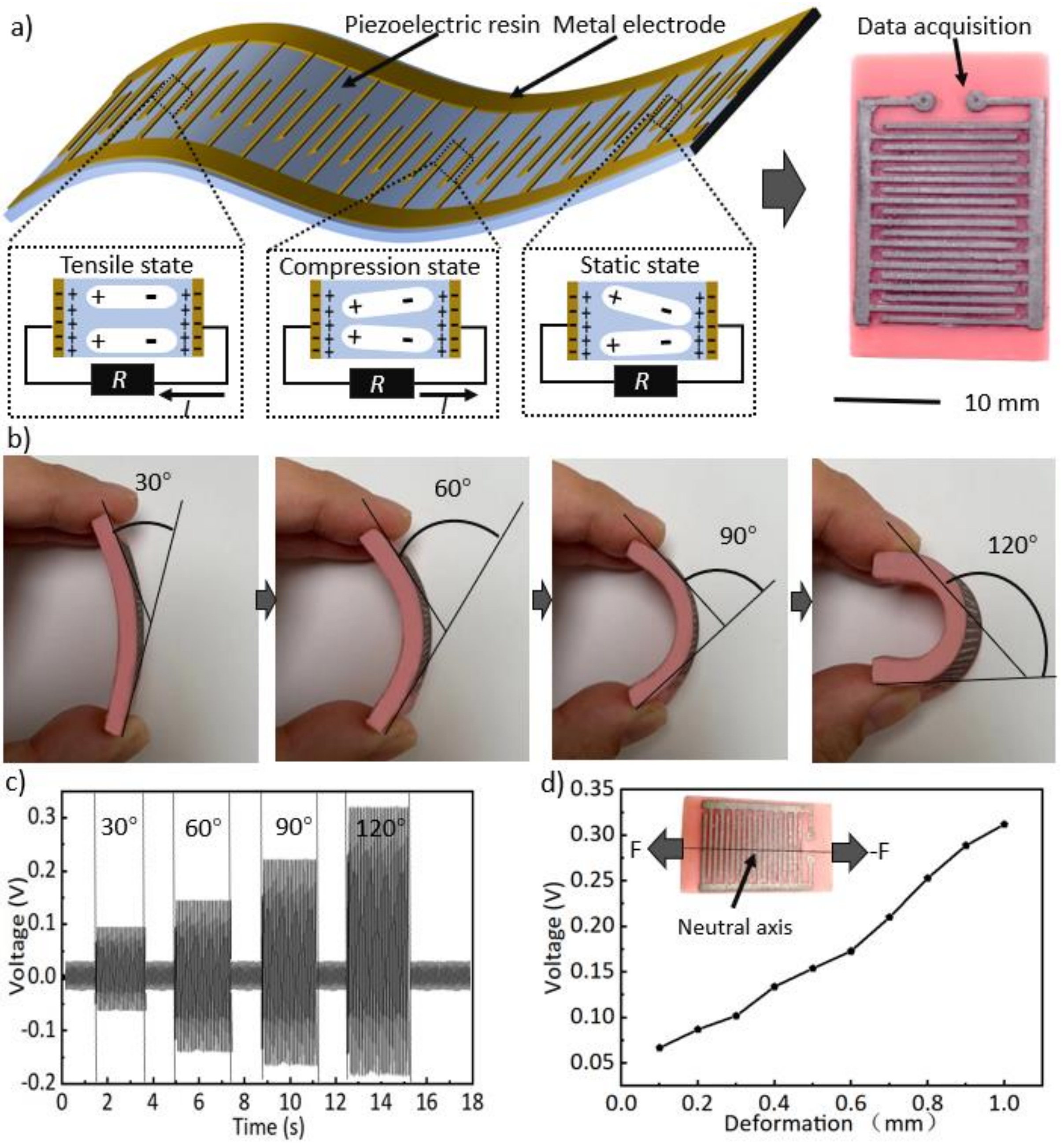

**Fig. 11. Principles and characteristics of 3D printed piezoelectric sensors.** a. The measurement principle of 3D printed piezoelectric sensors. As the sensor is stretched or squeezed the comb-shaped electrodes generate voltages. b. The state of a 3D printed piezoelectric sensor when bent at 30°, 60°, 90° and 120°. c. The voltage waveforms generated by the 3D printed piezoelectric sensor at bending angles of 30°, 60°, 90° and 120°. d. The voltage generated by the sensor in tension as a function of the deformation.

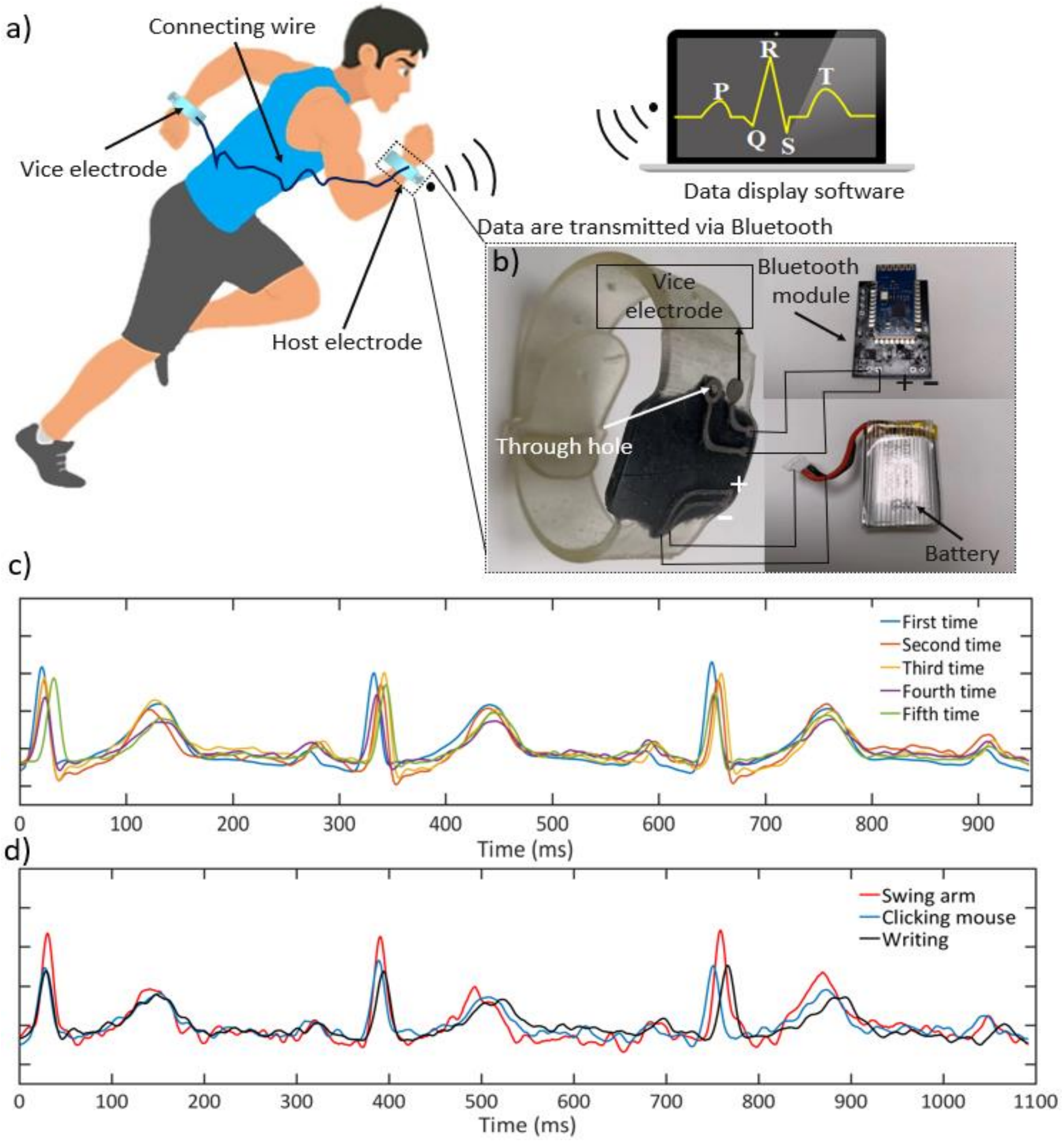

**Fig. 12. 3D printed integrated ECG electrode and its characteristics.** a. The working principle of the 3D printed integrated ECG electrode using a dual lead system in which metal electrodes in contact with the body generate ECG signals in real time that are sent to a computer for display via a Bluetooth module. b. Photographic images showing the device components. c. Five ECG signals acquired with the subject at rest, measured randomly. d. ECG signals with the subject swinging an arm, clicking a computer mouse, and writing.